\documentclass[sigconf]{acmart}
\AtBeginDocument{%
  \providecommand\BibTeX{{%
    \normalfont B\kern-0.5em{\scshape i\kern-0.25em b}\kern-0.8em\TeX}}}

\usepackage{booktabs}
\usepackage{array}
\usepackage{longtable}
\usepackage{graphicx}
\usepackage{caption}
\usepackage{subcaption}
\usepackage{amsmath} 
\usepackage{fvextra}
\usepackage{geometry}
\usepackage{enumitem}
\usepackage{float}
\usepackage{balance}
\copyrightyear{2024} 
\acmYear{2024} 
\setcopyright{acmlicensed}
\acmConference[MCHM '24] {Proceedings of the 1st International Workshop on Multimedia Computing for Health and Medicine}{October 28, 2024}{Melbourne, VIC, Australia.}
\acmBooktitle{Proceedings of the 1st International Workshop on Multimedia Computing for Health and Medicine (MCHM '24), October 28, 2024, Melbourne, VIC, Australia}
\acmISBN{979-8-4007-1195-4/24/10} 
\acmDOI{10.1145/3688868.3689200}

\begin{document}

\title{BURExtract-Llama: An LLM for Clinical Concept Extraction in Breast Ultrasound Reports}

% \title{BURExtract-Llama: A Large Language Model for Structured Clinical Concept Extraction in Breast Ultrasound Reports}

% \affiliation{%

% \author{Anonymous Authors}

\author{Yuxuan Chen$^\dagger$}
\affiliation{%
  \institution{New York University}
  % \streetaddress{1 Th{\o}rv{\"a}ld Circle}
  \city{New York}
  \country{USA}}
\email{yc7087@nyu.edu}

\author{Haoyan Yang$^\dagger$}
\affiliation{%
  \institution{New York University}
  % \streetaddress{1 Th{\o}rv{\"a}ld Circle}
  \city{New York}
  \country{USA}}
\email{hy2847@nyu.edu}

\author{Hengkai Pan}
\affiliation{%
  \institution{Carnegie Mellon University}
  % \streetaddress{1 Th{\o}rv{\"a}ld Circle}
  \city{Pittsburgh}
  \country{USA}}
\email{hengkaip@andrew.cmu.edu}

\author{Fardeen Siddiqui}
\affiliation{%
  \institution{NYU Langone Health}
  % \streetaddress{1 Th{\o}rv{\"a}ld Circle}
  \city{New York}
  \country{USA}}
\email{Fardeen.Siddiqui@nyulangone.org}

\author{Antonio Verdone}
\affiliation{%
  \institution{NYU Langone Health}
  % \streetaddress{1 Th{\o}rv{\"a}ld Circle}
  \city{New York}
  \country{USA}}
\email{Antonio.Verdone@nyulangone.org}

\author{Qingyang Zhang}
\affiliation{%
  \institution{NYU Shanghai}
  % \streetaddress{1 Th{\o}rv{\"a}ld Circle}
  \city{Shanghai}
  \country{China}}
\email{qz2208@nyu.edu}

\author{Sumit Chopra}
\affiliation{%
  \institution{NYU Langone Health}
  % \streetaddress{1 Th{\o}rv{\"a}ld Circle}
  \city{New York}
  \country{USA}}
\email{Sumit.Chopra@nyulangone.org}

\author{Chen Zhao}
\affiliation{%
  \institution{NYU Shanghai}
  % \streetaddress{1 Th{\o}rv{\"a}ld Circle}
  \city{Shanghai}
  \country{China}}
\email{cz1285@nyu.edu}

\author{Yiqiu Shen*}
\affiliation{%
  \institution{NYU Langone Health}
  % \streetaddress{1 Th{\o}rv{\"a}ld Circle}
  \city{New York}
  \country{USA}}
\email{Yiqiu.Shen@nyulangone.org}

%% You do not have to enter your paper ID

\renewcommand{\shortauthors}{Chen and Yang, et al.}

\begin{abstract}
Breast ultrasound is essential for detecting and diagnosing abnormalities, with radiology reports summarizing key findings like lesion characteristics and malignancy assessments. Extracting this critical information is challenging due to the unstructured nature of these reports, with varied linguistic styles and inconsistent formatting. While proprietary LLMs like GPT-4 are effective, they are costly and raise privacy concerns when handling protected health information. This study presents a pipeline for developing an in-house LLM to extract clinical information from radiology reports. We first use GPT-4 to create a small labeled dataset, then fine-tune a Llama3-8B model on it. Evaluated on clinician-annotated reports, our model achieves an average F1 score of 84.6\%, which is on par with GPT-4. Our findings demonstrate the feasibility of developing an in-house LLM that not only matches GPT-4's performance but also offers cost reductions and enhanced data privacy.
\end{abstract}






\maketitle

\def\thefootnote{$\dagger$}\footnotetext{\ These authors contributed equally to this work.}\def\thefootnote{\arabic{footnote}}

\def\thefootnote{*}\footnotetext{\ Corresponding author: Yiqiu Shen. }\def\thefootnote{\arabic{footnote}}

\section{Introduction}
Ultrasonography plays a crucial role in diagnosing breast pathology, with radiologists generating comprehensive reports following the BI-RADS guidelines to standardize lesion characterization ~\cite{flobbe2003additional, mendelson2013acr}. However, the large volume of data in these reports, combined with variations in reporting styles and clinical discrepancies, often leads to unstructured information that's challenging to extract systematically. Implementing a method to retrieve this clinical data could enhance Clinical Decision Support Systems (CDSS)  \cite{hak2022towards}, providing real-time alerts and recommendations, improving data accuracy, and benefiting both physicians and trainees by supporting case characterization and diagnostic decision tracking.

Commercial LLMs like GPT-4~\cite{openai2024gpt4} offer satisfactory accuracy in information retrieval but can be costly for large datasets and pose privacy risks with protected health information (PHI) in medical reports. To mitigate these issues, institutions often fine-tune open-source models like Llama 3~\cite{llama3modelcard}, which, while potentially less accurate, enhance privacy and reduce costs. However, developing in-house LLMs is challenging due to the need for large, high-quality annotated clinical datasets~\cite{navarro2023clinical} and the unstructured nature of radiology reports, which vary in style and format.

\begin{figure*}[t]
    \centering
    \includegraphics[width=\textwidth]{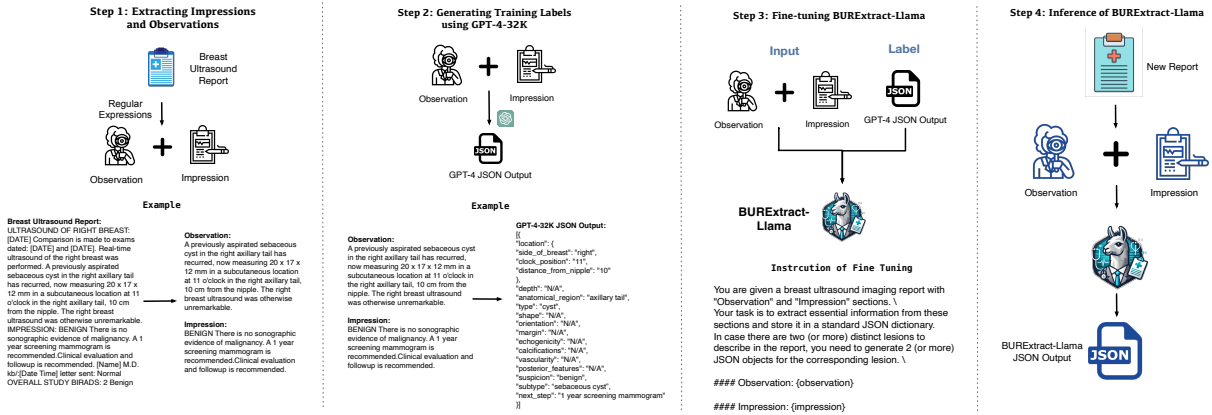}
    \caption{The pipeline for building and utilizing the BURExtract-Llama. Steps 1-3 cover the building process: extracting observations and impressions, generating training labels with GPT-4, and fine-tuning BURExtract-Llama. Step 4 shows how BURExtract-Llama infers from new reports and outputs structured JSON.}
    \label{fig1}
\end{figure*}

We aim to standardize a workflow for institutions to train an in-house LLM without the need for costly, large-scale manually annotated datasets. In this work, we present a pipeline for developing an in-house LLM to extract relevant clinical information from radiology reports. As depicted in Figure \ref{fig1}, our method begins by using a high-performance proprietary LLM, GPT-4, to create a small subset of labeled data. We then fine-tune an open-source LLM using this dataset. Evaluated on a subset manually annotated by clinicians, our model, named BURExtract-Llama, achieves high accuracy in extracting clinical information from \textbf{B}reast \textbf{U}ltrasound \textbf{R}eports, comparable to GPT-4.

\section{Related Works}
In clinical concept extraction, methodologies have undergone substantial evolution, progressing from early rule-based systems to deep learning approaches. Initially, the domain relied heavily on rule-based systems, such as the method proposed by Friedman et al., which employs a three-phase processing approach—parsing to identify text structure, regularization to standardize terms, and encoding to map terms to a controlled vocabulary \cite{friedman1994general}.

With technological growth, the introduction of machine learning models such as Conditional Random Fields \cite{esuli2013enhanced} and Support Vector Machines \cite{tang2013recognizing} shifted the field towards more dynamic analysis. Furthermore, the emergence of deep learning architectures, including Recurrent Neural Networks and Long Short-Term Memory models \cite{DBLP:journals/corr/abs-1808-03314}, has freed humans from manual feature engineering by employing distributed word representations \cite{peng2023clinical}, while effectively capturing the subtle nuances and complexities inherent in clinical text \cite{chalapathy2016bidirectional}.

The advent of the transformer  \cite{vaswani2017attention} marked another significant milestone. Transformer-based models such as BERT \cite{devlin2018bert}, ALBERT \cite{lan2019albert}, RoBERTa \cite{liu2019roberta}, and ELECTRA \cite{clark2020electra} have been explored for clinical concept extraction tasks \cite{yang2020clinical}. The incorporation of self-attention mechanisms in these models enhances the long-term dependencies management, thereby providing a more sophisticated tool for this task.

Recent advancements in LLMs, such as GPT-4 \cite{openai2024gpt4} and Llama 3 \cite{llama3modelcard}, have revolutionized the field of natural language processing. Their proficiency in tasks ranging from text summarization to question answering highlights their versatility. Consequently, our research aims to investigate the fine-tuning of LLMs for clinical concept extraction, demonstrating their capability to manage complex clinical text.

\section{Methods}
% [TODO Alex \& Haoyan: please give an example report and give the expected output.]
\subsection{Problem Formulation}
Formally, let \( \mathbf{x} \)  denote a breast ultrasound report, containing textual descriptions of one or multiple lesions. Our goal is to transform each report \( \mathbf{x} \) into a list of JSON dictionaries  \( \mathbf{y} = [y_{1}, y_{2}, \ldots, y_{n}] \), where each dictionary  \( y_{i} \) corresponds to a lesion described in \( \mathbf{x} \). Each dictionary \( y_{i} \) is defined as $y_{i} = \{ (k_1, v_1), (k_2, v_2), \ldots, (k_{m}, v_{m}) \}$, where \( k_1, k_2, \ldots, k_{m} \) are the keys representing lesion attributes, and \( v_1,v_2, \ldots, v_{m} \) are the corresponding values extracted from the report. In our study, there are 16 keys of interest, as shown in Table \ref{table1}.  Our objective is to train an LLM that processes \( \mathbf{x} \) to generate \( \mathbf{y} \).  Please refer to Appendix \ref{appendix-a} for an example of input and output.

\begin{table}[h]
\centering
\small
\caption{Keys and values of lesions}
\label{table1}
\small
\begin{tabular}{p{2.8cm}p{4cm}}
\hline
\textbf{Keys} & \textbf{Values} \\
\hline
k1=depth & Posterior, Middle, Anterior, N/A \\
\hline
k2=anatomical region & Retroareolar, Axillary Tail, Periareolar, Subareolar, Retropectoral, N/A \\
\hline
k3=lesion type & Nodule, Cyst, Mass, Lymph Node, Scar, Duct, Seroma, Post-Surgical Change, Post-Biopsy, N/A \\
\hline
k4=lesion shape & Oval, Round, Irregular, N/A \\
\hline
k5=orientation & Parallel, Non-Parallel, Other, N/A \\
\hline
k6=lesion margins & Circumscribed, Obscured, Angular, Microlobulated, Spiculated, Lobulated, Irregular, Septated, N/A \\
\hline
k7=echogenicity & Anechoic, Hyperechoic, Hypoechoic, Isoechoic, Heterogeneous, Solid, N/A  \\
\hline
k8=calcifications & Yes, No, N/A \\
\hline
k9=vascularity & Absent, Present, N/A \\
\hline
k10=posterior features & Enhancement, Shadowing, N/A \\
\hline
k11=lesion subtype & Abnormal Lymph Node, Simple Cyst, Complicated Cyst, Cyst with Debris, Reactive Lymph Node, Fat Necrosis, Sebaceous Cyst, Lipoma, Cyst Cluster, Focally Ectatic Duct with Debris, N/A \\
\hline
k12=next step & 1 Year Screening Mammogram, MRI Follow Up, 6 Months Follow-Up, 12 Months Follow-up, Fine Needle Aspiration, Ultrasound Guided Core Biopsy, Surgical Excision, N/A\\
\hline
k13=suspicion of malignancy & Low, Moderate, High, Benign, Probably Benign, Negative \\
\hline
k14=side of breast & Left, Right, N/A \\
\hline
k15=clock position & 1, 2, 3, 4, 5, 6, 7, 8, 9, 10, 11, 12, N/A \\
\hline
k16=distance from nipple & Numeric Value (cm) \\
\hline
\end{tabular}
\end{table}

% \begin{itemize}
%     \item \( k_1 = \text{"depth"} \)
%     \item \( k_2 = \text{"anatomical region"} \)
%     \item \( k_3 = \text{"type"} \)
%     \item \( k_4 = \text{"shape"} \)
%     \item \( k_5 = \text{"orientation"} \)
%     \item \( k_6 = \text{"margin"} \)
%     \item \( k_7 = \text{"echogenicity"} \)
%     \item \( k_8 = \text{"calcifications"} \)
%     \item \( k_9 = \text{"vascularity"} \)
%     \item \( k_{10} = \text{"posterior features"} \)
%     \item \( k_{11} = \text{"subtype"} \)
%     \item \( k_{12} = \text{"next step"} \)
%     \item \( k_{13} = \text{"suspicion"} \)
% \end{itemize}

\subsection{Pipeline for building in-house LLMs}
As illustrated in Figure \ref{fig1}, our pipeline consists of three steps: 1) extracting observations (also referred to as \textit{findings}) and impressions from reports; 2) generating training labels using GPT-4 \cite{openai2024gpt4}; 3) fine-tuning Llama3-8B using Q-LoRA~\cite{dettmers2024qlora}.

% and 4) post-processing the output from  Llama3-8B. Each step is detailed in the following subsections.

\subsubsection{Extract Observations and Impressions}
Figure 1 illustrates the components of a typical breast ultrasound report, which includes: 1) patient and examination details; 2) methodology of the ultrasound; 3) findings, detailing the notable features observed in the breast tissue; 4) impression, offering a concise summary, interpretation of findings, and recommendations for subsequent actions; and 5) disclosure. Unlike other sections whose information is available in a structured format, the observation and impression sections contain key descriptions of lesions in an unstructured manner. Consequently, we focus on extracting clinical information primarily from these two sections. We employ regular expressions to isolate the observation and impression sections, ensuring the input for the LLMs is in a clean format.

\subsubsection{Generate Training Labels using GPT-4-32K}
Our institute utilizes a HIPAA-compliant GPT-4 instance, allowing us to process medical records securely. Leveraging the in-context learning (ICL) capabilities~\cite{dong2022survey} of GPT-4-32K~\cite{openai2024gpt4}, we generate JSON labels to fine-tune the Llama-3 model. We design a prompt template that includes 7 carefully curated examples to illustrate the expected JSON format, ensuring clarity and consistency in the generated labels. For details on the prompt structure and examples, please refer to the appendix \ref{appendix-b}.

\subsubsection{Fine-Tuning}
We fine-tune Llama-3-8B by utilizing QLoRA~\cite{dettmers2024qlora} on pairs of report and JSON outputs generated by GPT-4. Fine-tuning allows Llama-3-8B to perform well on our specific task by adapting our dataset. We choose QLoRA, which is a quantized version of Low-Rank Adaptation (LoRA)~\cite{hu2021loralowrankadaptationlarge}, as it can significantly reduce memory and computational requirements.

First, we quantize the model weights as $\theta_q = Q(\theta)$. Instead of performing a full fine-tuning, we only update the two low-rank matrices $B$ and $A$ using backpropogration, as demonstrated in \eqref{eq5}. This approach reduces the size of the parameters that need to be updated from approximately 16GB to about 100MB compared to full fine-tuning, while maintaining model performance. The matrices $B$ and $A$ are updated through backpropagation based on the loss function defined in \eqref{eq3} and \eqref{eq4}. This ensures that the model minimizes the discrepancy between the predicted outputs and the actual outputs at the token level. 

In summary, the QLoRA method significantly reduces the number of parameters that need to be updated compared to a full fine-tuning, enhancing both memory and computational efficiency.
\begin{equation}
\label{eq5}
\Delta\theta_i \approx \frac{\alpha}{r} BA, \theta_i \leftarrow \theta_i + \Delta\theta_i
\end{equation}
where \( B \in \mathbb{R}^{d \times r} \) and \( A \in \mathbb{R}^{r \times k} \), with rank \( r \ll \min(d, k) \). $r$ represents the LoRA attention dimension and $\alpha$ is for LoRA scaling.

\begin{equation}
\label{eq3}
p_{\theta_q}(\mathbf{s}|\mathbf{x}) = \prod_{j=1}^{t} p_{\theta_q}(s_j|\mathbf{x}, \mathbf{s}_{<j})
\end{equation}

\begin{equation}
\label{eq4}
L(\theta_q) = - \mathbb{E}_{\mathbf{x}, \mathbf{s}}(\cdot|\mathbf{x}) \left[ \log p_{\theta_q}(\mathbf{s}|\mathbf{x}) \right]
\end{equation}
% where $x \in R$ as a input sequence \(\mathbf{x} = [x_1, \ldots, x_n]\) and $y \in L_{\text{true}} $ as a generated corresponding response \(\mathbf{y} = [y_1, \ldots, y_m]\).
where $\mathbf{s}$ is the output string, $s_{<j} = [s_{1},...,s_{j-1}]$ for $j = 2, . . . , t$, and $t$ is the length of string $\mathbf{s}$.

\subsection{Inference}
During the inference phase of BURExtract-Llama, a report undergoes processing to extract the observation and impression components. These elements are integrated with the fine-tuning instructions detailed in Step 3 of Figure \ref{fig1} to execute inference, producing a structured JSON output.

\section{Experiments}

\subsection{Dataset}
The dataset used in this study consist of 4000 breast ultrasound reports from NYU Langone. We divide this dataset into training (3600), validation (280), and test sets (120), with no overlapping reports. The training and validation sets are labeled using GPT-4. The test set is clinician-labeled to mitigate potential errors from automated labeling, ensuring accurate model evaluation.

\begin{table*}[htbp]
\centering
\small
\caption{Performance comparison of ICL of Llama3-Instruct-8B, BURExtract-Llama, and ICL of GPT-4 across 16 keys for precision, recall, and F1 score.}
\label{table2}
\begin{tabular}{l*{3}{>{\centering\arraybackslash}p{1.0cm}}*{3}{>{\centering\arraybackslash}p{1.0cm}}*{3}{>{\centering\arraybackslash}p{1.0cm}}}
\toprule
& \multicolumn{3}{c}{ICL of Llama3-Instruct-8B} & \multicolumn{3}{c}{BURExtract-Llama} & \multicolumn{3}{c}{ICL of GPT-4} \\
\cmidrule(lr){2-4} \cmidrule(lr){5-7} \cmidrule(lr){8-10}
& Recall & Precision & F1 & Recall & Precision & F1 & Recall & Precision & F1 \\
\midrule
depth & 0.798 & 0.906 & 0.849 & 0.853 & 0.935 & 0.892 & 0.839 & 0.934 & 0.884 \\
anatomical region & 0.757 & 0.859 & 0.805 & 0.807 & 0.884 & 0.844 & 0.780 & 0.867 & 0.821 \\
lesion type & 0.702 & 0.797 & 0.746 & 0.775 & 0.849 & 0.811 & 0.780 & 0.867 & 0.821 \\
lesion shape & 0.771 & 0.875 & 0.820 & 0.826 & 0.905 & 0.863 & 0.835 & 0.929 & 0.879 \\
orientation & 0.803 & 0.911 & 0.854 & 0.858 & 0.940 & 0.897 & 0.858 & 0.954 & 0.903 \\
lesion margins & 0.761 & 0.865 & 0.810 & 0.830 & 0.910 & 0.868 & 0.826 & 0.918 & 0.870 \\
echogenicity & 0.743 & 0.844 & 0.790 & 0.794 & 0.869 & 0.830 & 0.798 & 0.888 & 0.841 \\
calcifications & 0.812 & 0.922 & 0.863 & 0.867 & 0.950 & 0.906 & 0.849 & 0.944 & 0.894 \\
vascularity & 0.789 & 0.896 & 0.839 & 0.849 & 0.930 & 0.887 & 0.835 & 0.929 & 0.879 \\
posterior features & 0.789 & 0.896 & 0.839 & 0.862 & 0.945 & 0.902 & 0.853 & 0.949 & 0.899 \\
lesion subtype & 0.693 & 0.786 & 0.737 & 0.784 & 0.859 & 0.820 & 0.766 & 0.852 & 0.807 \\
next step & 0.674 & 0.766 & 0.717 & 0.743 & 0.814 & 0.777 & 0.748 & 0.832 & 0.787 \\
suspicion of malignancy& 0.564 & 0.641 & 0.600 & 0.615 & 0.673 & 0.643 & 0.638 & 0.709 & 0.671 \\
side of breast & 0.817 & 0.927 & 0.868 & 0.872 & 0.955 & 0.911 & 0.858 & 0.954 & 0.903 \\
clock position & 0.734 & 0.833 & 0.780 & 0.803 & 0.879 & 0.839 & 0.798 & 0.888 & 0.841 \\
distance from nipple & 0.743 & 0.844 & 0.790 & 0.807 & 0.884 & 0.844 & 0.803 & 0.893 & 0.845 \\
\midrule
Average & 0.747 & 0.848 & 0.794 & 0.809 & 0.886 & 0.846 & 0.804 & 0.894 & 0.847 \\
\bottomrule
\end{tabular}
\label{tab2}
\end{table*}

\subsection{Training Detail}
We include all hyper-parameters in Appendix \ref{appendix-c}. During the validation, we focus on the \textit{number of training epochs} and the \textit{LoRA attention dimension} to select the best model. As detailed in Appendix \ref{appendix-d}, epoch 4 and a LoRA attention dimension of 64 yielded the best performance. Our best model was trained on an Nvidia A100 GPU for 1.5 hours.

\subsection{Evaluation Metrics}
We employ two primary categories of evaluation metrics: per report matching and per key matching. \textbf{Per Report Matching} includes three metrics:

% \noindent \textbf{JSONable Accuracy}: The percentage of the LLM's outputs that can be converted into a valid list of dictionaries. 

% \noindent \textbf{Exact Matching (EM) Accuracy}: The percentage of the LLM's outputs, after converted to JSON dictionaries, can exactly match the ground truth for all 16 keys \( \{ k1, k2 \dots k16 \} \). 

% \noindent \textbf{Close Domain Matching (CDM) Accuracy}: The percentage of the LLM's outputs that match the ground truth based on a subset of keys \( \{ k1, k2 \dots k10 \} \) which contains categorical values that provides critical information for diagnosis.
\begin{itemize}[itemsep=0pt,parsep=0pt,leftmargin=*]
    \item \textbf{JSONable Accuracy}: The percentage of the LLM's outputs that can be converted into a valid list of dictionaries.
    \item \textbf{Exact Matching (EM) Accuracy}: The percentage of the LLM's outputs, after converted to JSON, that exactly match the ground truth for all 16 keys \( \{ k1, k2 \dots k16 \} \).
    % \item \textbf{Close Domain Matching (CDM) Accuracy}: The percentage of the LLM's outputs that match the ground truth based on a subset of keys \( \{ k1, k2 \dots k10 \} \) which contains categorical values that provides critical information for diagnosis.
    \item \textbf{Close Domain Matching (CDM) Accuracy}: The percentage of the LLM outputs that match the ground truth for a subset of keys \( \{ k1, k2 \dots k10 \} \) containing categorical values critical for diagnosis.
\end{itemize}

\textbf{Per Key Matching} calculates recall, precision, and F1 score for each of 16 keys. The details for both per key matching and per report matching, including their mathematical definitions, are provided in Appendix \ref{appendix-e}.

\subsection{Results}
Table \ref{table2} provides a detailed comparison of precision, recall, and F1 scores for each key. The average F1 score difference between BURExtract-Llama and GPT-4's ICL is within 0.1\%. Notably, BURExtract-Llama outperforms GPT-4 in keys like "depth," "anatomical region," and "posterior features," with a maximum difference of 2.3\% in "anatomical region".

As shown in Table \ref{table3}, BURExtract-Llama achieves 100\% structured output for the JSONable accuracy, demonstrating its ability to follow the prompt instructions and produce the desired output format. Our BURExtract-Llama outperforms Llama3-8B with ICL by 12.5\% in EM and 10.0\% in CDM, highlighting the benefits of fine-tuning. It matches GPT-4 in EM and is only 0.9\% behind in CDM, proving BURExtract-Llama to be a viable alternative to GPT-4. Additionally, our BURExtract-Llama can infer a new report in approximately 2 seconds with vllm \cite{kwon2023efficientmemorymanagementlarge}, demonstrating its low latency.

\begin{table}[htbp]
    \centering
    \small
    \caption{Comparison of Results: ICL of Llama3-Instruct-8B, BURExtract-Llama, and ICL of GPT-4.}
    \label{table3}
    \begin{tabular}{p{1.0cm}p{1.7cm}p{1.7cm}p{1.7cm}}
        \toprule
        & ICL of Llama3-Instruct-8B & BURExtract-Llama & ICL of GPT-4 \\
        \midrule
        JSONable & \textbf{1.000} & \textbf{1.000} & \textbf{1.000} \\
        EM & 0.333 & \textbf{0.458} & \textbf{0.458} \\
        CDM & 0.583 & 0.683 & \textbf{0.692} \\
        \bottomrule
    \end{tabular}

    \label{tab1}
\end{table}

\subsection{Error Analysis}
We conduct an error analysis to identify instances where BURExtract-Llama failed. While the detailed error cases are listed in the appendix \ref{appendix-f}, we present a summary here:

% \noindent \textbf{Missing Lesion}:
% The model sometimes fails to cover all lesions within a report, as reflected by the lower recall rate compared to precision in Table 2.

% \noindent \textbf{Lesion Attribute Confusion}: The model occasionally misattributes the values of one lesion to another, leading to incorrect associations.

% \noindent \textbf{Handling "N/A"}: The ground truth is marked as "N/A" when a report doesn't mention information for an attribute. Our model sometimes exhibits inconsistency in handling these cases, predicting "N/A" for attributes with actual values. This issue likely arises from the high frequency of "N/A" values in the training set for some attributes, as detailed in Appendix \ref{appendix-g}.

\begin{itemize}[itemsep=0pt,parsep=0pt,leftmargin=*]  
    \item \textbf{Missing Lesion}:
    The model sometimes fails to cover all lesions within a report, as reflected by the lower recall rate compared to precision in Table 2.

     \item \textbf{Lesion Attribute Confusion}: The model occasionally misattributes the values of one lesion to another, leading to incorrect associations.

     \item \textbf{Handling "N/A"}: The ground truth is marked as "N/A" when a report doesn't mention information for an attribute. Our model sometimes exhibits inconsistency in handling these cases, predicting "N/A" for attributes with actual values. This issue likely arises from the high frequency of "N/A" values in the training set for some attributes, as detailed in Appendix \ref{appendix-g}.

     % \item \textbf{Handling "N/A"}: The model sometimes predicts "N/A" for attributes with actual values, likely due to the high frequency of "N/A" in the training set for some attributes, as detailed in Appendix \ref{appendix-g}.
    
\end{itemize}

\section{Conclusion}
This study presents a workflow of building in-house LLM to extract relevant clinical information from radiology reports. We utilize GPT-4 to label a small dataset, then fine-tune Llama3-8B on it. The fine-tuned model, BURExtract-Llama, demonstrates performance comparable to GPT-4. Still, we recognize several limitations. First, we acknowledge that the training labels generated by GPT-4 might be flawed. Future research could explore methods to manage noisy labels. The second limitation is the lack of external validation. Since the writing style of reports from a single hospital tends to be consistent, it is uncertain how the model would perform on data from institutions with different writing styles. Future research should also incorporate an external test set to evaluate the generalizability of the BURExtract-Llama.

\begin{acks}
This work was supported in part by the Murray J. Berenson, MD Grant in Medical Education Research from the NYU Program for Medical Education Innovations and Research.
\end{acks}

\bibliographystyle{ACM-Reference-Format}
\balance
\bibliography{sample-base}

\appendix
\onecolumn

\DefineVerbatimEnvironment{MyVerbatim}{Verbatim}{
  fontsize=\small,
  breaklines=true,
  breakindent=0pt, % Set to no indent on line breaks
  breaksymbolleft={},
  commandchars=\\\{\},
  breakanywhere=true
}

\section{Example input and output of LLM}
\label{appendix-a}
\begin{MyVerbatim}
#### INPUT:
"""
Observation:
At the right 9:00 axis, 1 cm from the nipple, there is a 0.4 cm simple cyst. At the left 6:00 retroareolar location, there is a 0.6 x 0.4 x 0.6 cm hypoechoic mass, probably benign may represent a debris-filled cyst and corresponding to the previous mass annotated at the "left nipple posterior" location on exam dated MM/DD/YYYY. No other suspicious cystic or solid masses identified in either breast.

Impression:
PROBABLY BENIGN - FOLLOW-UP RECOMMENDED Left 6:00 retroareolar 0.6 cm hypoechoic mass, probably benign cyst with debris. Recommend follow-up ultrasound in 6 months to assess stability. I personally discussed the findings and recommendations with the patient.
"""

### OUTPUT:
"""
[\{
"location": \{
"side_of_breast": "right",
"clock_position": "9",
"distance_from_nipple": "1"
\},
"depth": "N/A",
"anatomical_region": "N/A",
"type": "cyst",
"shape": "N/A",
"orientation": "N/A",
"margin": "N/A",
"echogenicity": "N/A",
"calcifications": "N/A",
"vascularity": "N/A",
"posterior_features": "N/A",
"suspicion": "N/A",
"subtype": "N/A",
"next_step": "N/A"
\},
\{
"location": \{
"side_of_breast": "left",
"clock_position": "6",
"distance_from_nipple": "N/A"
\},
"depth": "N/A",
"anatomical_region": "retroareolar",
"type": "mass",
"shape": "N/A",
"orientation": "N/A",
"margin": "N/A",
"echogenicity": "hypoechoic",
"calcifications": "N/A",
"vascularity": "N/A",
"posterior_features": "N/A",
"suspicion": "probably benign",
"subtype": "cyst with debris",
"next_step": "follow-up ultrasound in 6 months"
\}]
"""
\end{MyVerbatim}

\section{Prompt Template for Generating JSON Lables}
\label{appendix-b}
% \lstset{
%   basicstyle=\ttfamily\small,
%   breaklines=false,
%   frame=none,
%   showstringspaces=false,
%   tabsize=1,
%   breakatwhitespace=true,
%   keepspaces=true,
%   columns=fullflexible,
%   xleftmargin=0pt,
%   resetmargins=true
% }

\begin{MyVerbatim}
You are given a breast ultrasound imaging report with "Observation" and "Impression" sections. Your task is to extract essential information from these sections and store it in a standard JSON dictionary. The JSON dictionary should contain the following keys:

"""
- location: (Side of breast, Clock Position, distance from the nipple)
- depth: the depth of the lesion (e.g., posterior, middle, anterior)
- anatomical_region: anotomical region of the breast (e.g., retroareolar, axillary_tail, periareolar, subareolar, retropectoral, N/A)
- type: Type of lesion in the observation (e.g., Mass, Cyst, Fibrocystic Tissue, Other Benign, Scar, Lymph Node, Duct, Fluid collection, Postbiopsy, Implant, Lump, N/A).
- shape: Shape of the lesion (e.g., Oval, Round, Irregular, Indeterminant, Lobulated, N/A).
- orientation: Orientation of the lesion (e.g., Parallel, Non-parallel, Other, N/A).
- margin: Margin characteristics (e.g., Circumscribed, Angular, Microlobulated, Indistinct, Ill-defined, Spiculated, Lobulated, Irregular, Septated, N/A).
- echogenicity: Echogenicity of the lesion (e.g., Anechoic, Hyperechoic, Hypoechoic, Isoechoic, Heterogeneous, Solid, N/A).
- calcifications: Presence of calcifications (e.g., Yes, No, N/A).
- vascularity: describe the condition of vascularity
- posterior_features: the characteristics of the tissue immediately behind (posterior to) the lesion (e.g., Enhancement, Shadowing, Complex, No, N/A)
- Suspicion: the suspicion of malignancy presented in the impression (e.g., bengin, probably_benign, low_suspicion_of_malignancy, moderate_suspicion_of_maligancy, high_suspicion_of_maligancy, N/A)
- subtype: The diagnosis in the Impression part
- next_step: the recommended next step to do in the impression

"""
### Template of Example:
"""

#### Input Report:

--- 

#### Observation:
[Observation text]

#### Impression:
[Impression text]

--- 

#### Expected output format:
```
[\{
  "location": \{
    "side_of_breast": "content",
    "clock_position": "content",
    "distance_from_nipple": "content"
  \},
  "depth": "content",
  "anatomical_region": "content",
  "type": "content",
  "shape": "content",
  "orientation": "content",
  "margin": "content",
  "echogenicity": "content",
  "calcifications": "content",
  "vascularity": "content",
  "posterior_features": "content",
  "suspicion": "content",
  "subtype": "content",
  "next_step": "content"
\}]
```

"""

### Examples
### Example 1

"""

#### Input Report:

---

#### Observation:
[Example Observation Input 1]
#### Impression:
[Example Impression Input 1]
---

#### Expected output format:
```
[Example Output 1]
```
"""

### Example 2

"""

#### Input Report:

---

#### Observation:
[Example Observation Input 2]
#### Impression:
[Example Impression Input 2]
---

#### Expected output format:
```
[Example Output 2]
```
"""

### Example 3

"""

#### Input Report:

---

#### Observation:
[Example Observation Input 3]
#### Impression:
[Example Impression Input 3]
---

#### Expected output format:
```
[Example Output 3]
```
"""

### Example 4

"""

#### Input Report:

---

#### Observation:
[Example Observation Input 4]
#### Impression:
[Example Impression Input 4]
---

#### Expected output format:
```
[Example Output 4]
```
"""

### Example 5

"""

#### Input Report:

---

#### Observation:
[Example Observation Input 5]
#### Impression:
[Example Impression Input 5]
---

#### Expected output format:
```
[Example Output 5]
```
"""

### Example 6

"""

#### Input Report:

---

#### Observation:
[Example Observation Input 6]
#### Impression:
[Example Impression Input 6]
---

#### Expected output format:
```
[Example Output 6]
```
"""

### Example 7

"""

#### Input Report:

---

#### Observation:
[Example Observation Input 7]
#### Impression:
[Example Impression Input 7]
---

#### Expected output format:
```
[Example Output 7]
```
"""


## Instructions
1. Use the provided examples to understand the format.
2. Extract the relevant information from the given "Observation" and "Impression" sections.
3. Construct the JSON dictionary with the extracted information


## Additional Points to consider:

1. In case there are two (or more) distinct lesions to describe in the report, you need to generate 2 (or more) JSON objects for the corresponding lesion.
2. If the information for a key is not found in the report, just leave it there as "N/A."
3. The above explanation of the JSON keys may not exhaust all the classes. Add new classes as needed if they are encountered in the reports.
4. If the 'Impression' section is not found, you can leave the 'subtype', 'suspicion', and 'next_step' fields as 'N/A'."
5. The output should be a list of python dictionary
6. If the report does not explicitly mention the presence of the calcification, leave it as N/A
7. If the report does not explicitly mention the type of the lesion, leave it as N/A
8. The "distance_from_nipple" within the "location" key in the output should always be a numerical value and should not be confused with the depth field. The value of distance_from_nipple should never contain terms like 'anterior', 'middle', or 'posterior'. 
9. If the report indicates a prior lesion has not recurred, do not generate the JSON for that lesion.


# The input is as follows:
"""
#### Input Report:

---

#### Observation:
[Observation Input]

#### Impression:
[Impression Input]
---

"""
\end{MyVerbatim}

% Continue with other examples as needed...

\newpage
\section{Hyper-parameters}
\label{appendix-c}
\begin{table}[h]
\centering
\begin{tabular}{>{\centering\arraybackslash}p{0.3\linewidth} >{\centering\arraybackslash}p{0.3\linewidth} >{\centering\arraybackslash}p{0.3\linewidth}}
\hline
QLoRA Parameters & Bitsandbytes Parameters & Training Parameters \\
\hline
LoRA attention dimension: 64 & Use 4-bit precision: True & Number of training epochs: 4\\
Alpha parameter for LoRA scaling: 16 & Compute dtype for 4-bit: float16 & Initial learning rate: 2e-4  \\
Dropout probability for LoRA: 0.1 & Quantization type: nf4 & Learning rate schedule: constant \\
&  Use nested quantization: False & Optimizer: Adam  \\
&  & Weight decay: 0.001 \\
&  & Warmup ratio: 0.03 \\
&  & Enable bf16 training: True \\
&  & Per device train batch size: 4 \\
&  & Per device eval batch size: 4  \\
&  & Gradient accumulation steps: 1 \\
&  & Maximum gradient norm: 0.3 \\
&  & Packing: False \\
\hline
\end{tabular}
\end{table}

\section{Hyper-parameter optimization results}
\label{appendix-d}
The details for the four metrics in the legend can be found in section 4.3.
\begin{figure}[htbp]
    \centering
    \begin{subfigure}[b]{0.70\textwidth}
        \centering
        \includegraphics[width=\textwidth]{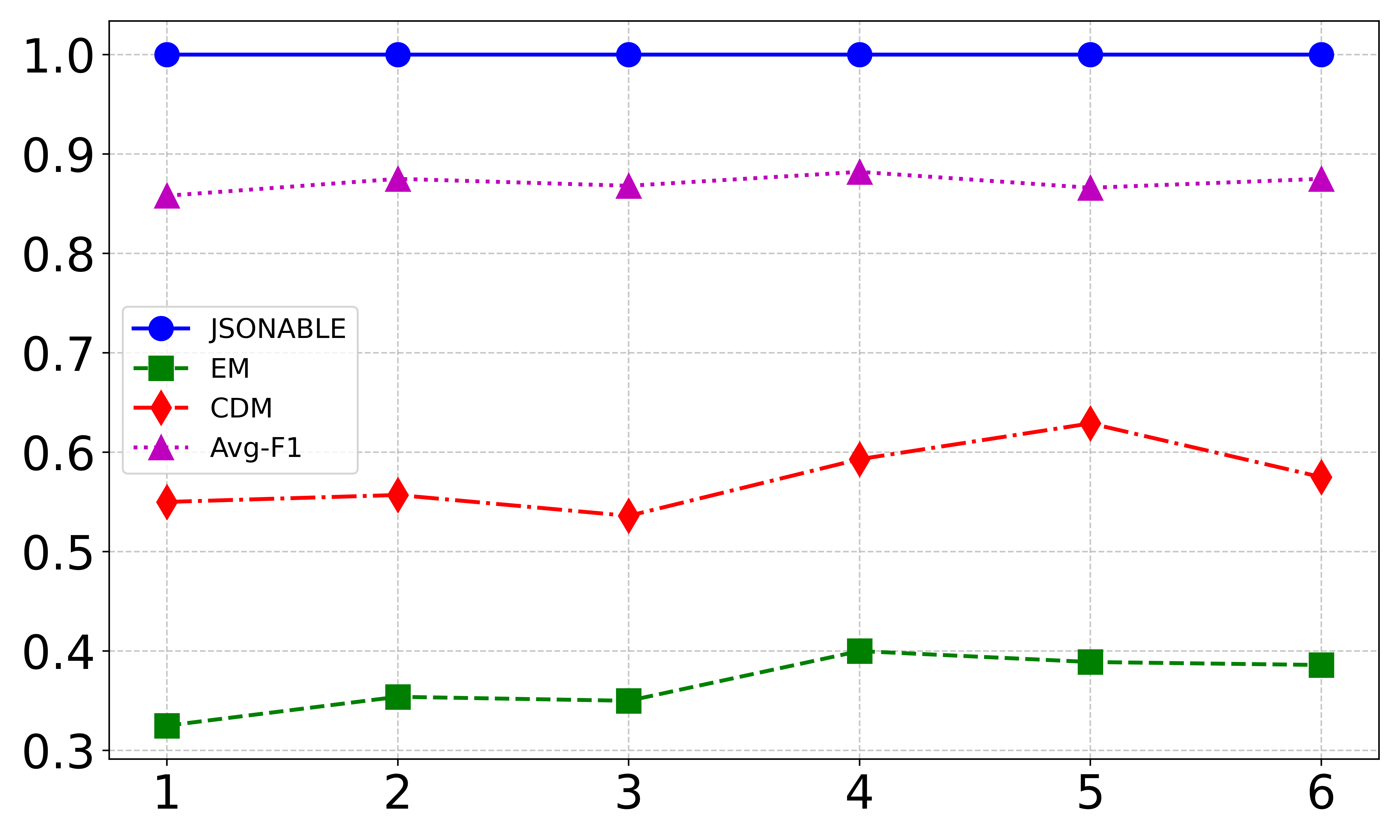}
        \caption{Epochs}
    \end{subfigure}
    \begin{subfigure}[b]{0.70\textwidth}
        \centering
        \includegraphics[width=\textwidth]{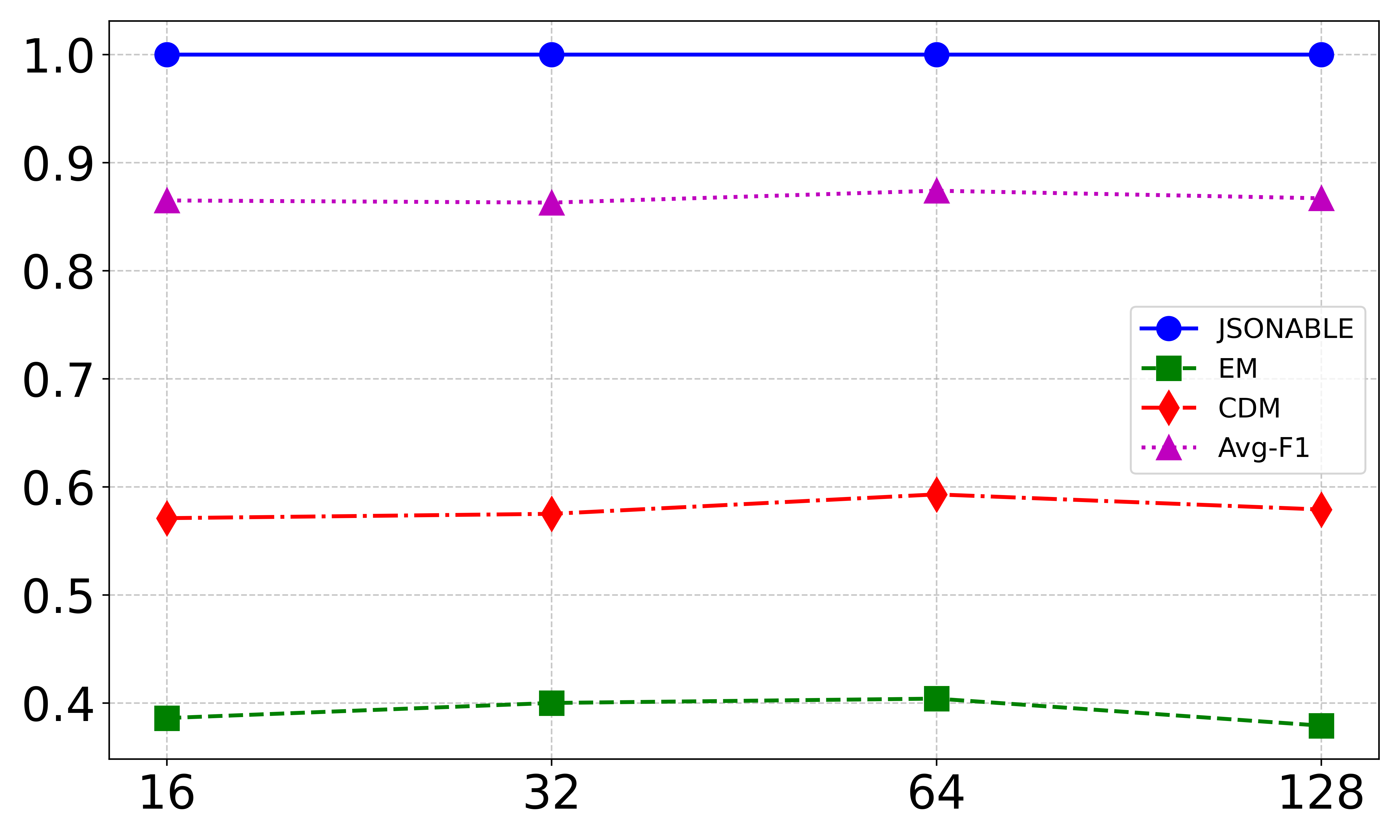}
        \caption{Attention Dimension}
    \end{subfigure}
    % \label{fig2}
\end{figure}

\section{Details of Metrics}
\label{appendix-e}
\textbf{JSONable Accuracy}: As shown in Equation \eqref{eq:jsonable}, a report \( \mathbf{x} \) is considered jsonable if the model’s prediction \( \mathbf{y'} \) is a valid list of JSON dictionaries:
\begin{equation}
\label{eq:jsonable}
\text{JSONable}(\mathbf{x}) = \text{IsValidJSONList}(\mathbf{y'})
\end{equation}
The JSONable accuracy is then calculated as the proportion of reports that are jsonable, as described in Equation \eqref{eq:jsonable_acc}:
\begin{equation}
\label{eq:jsonable_acc}
\text{JSONable\_Acc} = \frac{\sum_{\mathbf{x} \in X} \mathbb{I}(\text{JSONable}(\mathbf{x}))}{|\mathbf{X}|}
\end{equation}
where \( \mathbf{X} \) comprises all the reports, and \( \mathbb{I} \) is the indicator function that returns 1 if the condition is true and 0 otherwise.

\noindent \textbf{CDM Accuracy}: Equation \eqref{eq:lenmatch} defines the first condition for CDM, which checks whether the lengths of the predicted list \( \mathbf{y'}\) and true list \( \mathbf{y} \) are equal:
\begin{equation}
\label{eq:lenmatch}
\text{LenMatch}(\mathbf{x}) = (|\mathbf{y'}| = |\mathbf{y}|)
\end{equation}
where the lists for both \( \mathbf{y'} \) and \( \mathbf{y} \) are sorted by key \( k_{14}\) side of breast before comparison to maintain the correct order of the lesions.

Equation \eqref{eq:cdm_cond} defines the second condition for CDM. This condition requires that for each report \( \mathbf{x} \), the values of the keys in \( K_{\text{close}} \) in the prediction list \( y'_{i}[k] \) must exactly match the corresponding values in the ground truth list \( y_{i}[k] \) for every element \( i \) from 1 to \( n \), where \( K_{\text{close}} \) represents the set of keys \( \{k_1, k_2, \ldots, k_{10}\} \).

\begin{equation}
\label{eq:cdm_cond}
\text{CDM\_Cond}(\mathbf{x}) = \bigwedge_{i=1}^{n} \bigwedge_{k \in K_{\text{close}}} (y'_{i}[k] = y_{i}[k])
\end{equation}
As shown in Equation \eqref{eq:cdm}, for a report \( \mathbf{x} \) to be considered a Close Domain Match, both conditions must be met:
\begin{equation}
\label{eq:cdm}
\text{CDM}(\mathbf{x}) = \text{LenMatch}(\mathbf{x}) \land \text{CDM\_Cond}(\mathbf{x})
\end{equation}
The Close Domain Matching accuracy is calculated as the proportion of reports that are close domain matches, as described in Equation \eqref{eq:cdm_acc}:
\begin{equation}
\label{eq:cdm_acc}
\text{CDM\_Acc} = \frac{\sum_{\mathbf{x} \in \mathbf{X}} \mathbb{I}(\text{CDM}(\mathbf{x}))}{|\mathbf{X}|}
\end{equation}

\noindent \textbf{EM Accuracy}: Equation \eqref{eq:em_cond} defines the conditions for Exact Matching (EM), where \( K_{\text{exact}} \) represents the set of keys \( \{k_1, k_2, \ldots, k_{16}\} \):
\begin{equation}
\label{eq:em_cond}
\text{EM\_Cond}(\mathbf{x}) = \bigwedge_{i=1}^{n} \bigwedge_{k \in K_{\text{exact}}} (\mathbf{y'_{i}[k]} = \mathbf{y_{i}[k]})
\end{equation}
As shown in Equation \eqref{eq:em}, for a report \( \mathbf{x} \) to be considered an Exact Match, both conditions must be satisfied:
\begin{equation}
\label{eq:em}
\text{EM}(\mathbf{x}) = \text{LenMatch}(\mathbf{x}) \land \text{EM\_Cond}(\mathbf{x})
\end{equation}
The Exact Matching accuracy is calculated as the proportion of reports that are exact matches, as described in Equation \eqref{eq:em_acc}:
\begin{equation}
\label{eq:em_acc}
\text{EM\_Acc} = \frac{\sum_{\mathbf{x} \in \mathbf{X}} \mathbb{I}(\text{EM}(\mathbf{x}))}{|\mathbf{X}|}
\end{equation}

To evaluate the model's performance on specific key attributes, we calculate recall, precision, and F1 score for each key \( k \). Given that the lengths of the prediction list and the ground truth list may vary, simply sorting the lists and iterating over them for comparison may not be fair, as the order can still be incorrect. Therefore, we split both the prediction list and the ground truth list into three separate lists based on the side of the breast: left, right and N/A. Comparisons are then made between the corresponding lists, \( \mathbf{y'_{side}}\) and \( \mathbf{y_{side}}\). The formal definitions of these evaluation metrics are as follows:

\noindent \textbf{Recall}: Measures the model’s ability to identify all relevant instances of a specific key \( k \). Recall is quantified as the ratio of correctly predicted instances to the total instances present in the ground truth. This is computed using equations \eqref{eq:minlen} - \eqref{eq:recall} 

\begin{equation}
\label{eq:minlen}
\text{MinLen}(\mathbf{x}, \text{side}) = \min(|\mathbf{y_{\text{side}}}|, |\mathbf{y'_{\text{side}}}|)
\end{equation}

\begin{equation}
\label{eq:recall_side}
\text{CountMatch}(\mathbf{x}, k, \text{side}) = \sum_{i=1}^{\text{MinLen}(\mathbf{x},\text{side})} \mathbb{I}(y_{\text{side},i}[k] = y'_{\text{side},i}[k])
\end{equation}

\begin{equation}
\label{eq:recall}
\text{Recall}(k) = \frac{\sum_{\mathbf{x} \in \mathbf{X}} \left(\text{CountMatch}(\mathbf{x}, k, \text{left}) + \text{CountMatch}(\mathbf{x}, k, \text{right}) + \text{CountMatch}(\mathbf{x}, k, \text{N/A})\right)}
{\sum_{\mathbf{y} \in \mathbf{Y}}|\mathbf{y}|}
\end{equation}
where $\mathbf{Y}$ is the set that contains all the ground truth.
% (|\mathbf{y_{\text{left}}}| + |\mathbf{y_{\text{right}}}| + |\mathbf{y_{\text{N/A}}|)}}

\noindent \textbf{Precision}: Evaluates the proportion of true positive predictions among all predictive instances on key \(k\). Precision is defined as the ratio of correctly predicted instances to the total number of predicted instances. This is calculated in equation \eqref{eq:precision}:

\begin{equation}
\label{eq:precision}
\text{Precision}(k) = \frac{\sum_{\mathbf{x} \in \mathbf{X}} \left(\text{CountMatch}(\mathbf{x}, k, \text{left}) + \text{CountMatch}(\mathbf{x}, k, \text{right}) + \text{CountMatch}(\mathbf{x}, k, \text{N/A})\right)}{\sum_{\mathbf{y'} \in \mathbf{Y'}}|\mathbf{y'}|}
\end{equation}
where $\mathbf{Y'}$ is the set that contains all the prediction list.

\noindent \textbf{F1 Score}: Provides a single metric that balances both recall and precision. The F1 score is the harmonic mean of recall and precision and is defined as \eqref{eq:f1}:

\begin{equation}
\label{eq:f1}
\text{F1}(k) = 2 \cdot \frac{\text{Precision}(k) \cdot \text{Recall}(k)}{\text{Precision}(k) + \text{Recall}(k)}
\end{equation}

% \textbf{Recall}: Measures the model’s ability to identify all relevant instances of a specific key \( k \).Recall is quantified as the ratio of correctly predicted instances to the total instances present in the ground truth. This is computed using equations \ref{eq:min_length} and \ref{eq:recall}:
% \begin{equation}
% \label{eq:min_length}
% \text{MinLen}(r) = \min(|l'_{r}|, |l_{r}|)
% \end{equation}
% \begin{equation}
% \label{eq:recall}
% \text{Recall}(k) = \frac{\sum_{r \in R} \sum_{i=1}^{\text{MinLen}(r)} \mathbb{I}(j_{r,i}[k] = j'_{r,i}[k])}{\sum_{r \in R} |l_{r}|}
% \end{equation}

% \textbf{Precision}: Assesses the model's accuracy in predicting instances of a specific key \( k \). Precision is defined as the ratio of correctly predicted instances to the total number of predicted instances, as shown in equation \ref{eq:precision}.

% \begin{equation}
% \label{eq:precision}
% \text{Precision}(k) = \frac{\sum_{r \in R} \sum_{i=1}^{\text{MinLen}(r)} \mathbb{I}(j'_{r,i}[k] = j_{r,i}[k])}{\sum_{r \in R} |l'_{r}|}
% \end{equation}

% \textbf{F1 Score}: Provides a single metric that balances both recall and precision. The F1 score is the harmonic mean of recall and precision and is defined in equation \ref{eq:f1}:
% \begin{equation}
% \label{eq:f1}
% \text{F1}(k) = 2 \cdot \frac{\text{Precision}(k) \cdot \text{Recall}(k)}{\text{Precision}(k) + \text{Recall}(k)}
% \end{equation}

\section{Three different kinds of Error Cases}
\label{appendix-f}
% \noindent 1. The prediction has different value with the truth
% \begin{MyVerbatim}
% prediction: retroareolar
% truth: periarerolar

% prediction: axilla
% truth: n/a

% prediction: n/a
% truth: outer

% prediction: n/a
% truth: marginated

% prediction: n/a
% truth: irregular
% \end{MyVerbatim}

% \noindent 2. The length of the prediction list is longer than the ground truth list. So model gives a prediction which doesn't exist in the ground truth. 
% \begin{MyVerbatim}
% prediction: n/a
% [no truth]

% \end{MyVerbatim}

% \noindent 3. The length of the prediction list is shorter than the ground truth list. So the ground truth will give a label which doesn't exist in model's prediction. 
% \begin{MyVerbatim}
% [no prediction]
% truth: n/a
% \end{MyVerbatim}

% \noindent}
\textbf{1. Missing Lesion}
\begin{MyVerbatim}
Observation: The study demonstrates heterogeneous background echotexture. Again noted is an area of postsurgical distortion with a 9 x 9 x 4 mm seroma in the upper outer left breast. There is a 3 x 2 mm cyst in the left breast 3:00 N9 location. No suspicious abnormalities were seen sonographically in the left.

Impression: No Impression

\textbf{Prediction:}
[\{'location': \{'side_of_breast': 'left',
   'clock_position': '3',
   'distance_from_nipple': '9'\},
   'depth': 'n/a',
   'anatomical_region': 'n/a',
   'type': 'cyst',
   'shape': 'n/a',
   'orientation': 'n/a',
   'margin': 'n/a',
   'echogenicity': 'n/a',
   'calcifications': 'n/a',
   'vascularity': 'n/a',
   'posterior_features': 'n/a',
   'suspicion': 'n/a',
   'subtype': 'n/a',
   'next_step': 'n/a'\}]
\textbf{Ground Truth:}
[\{'location': \{'side_of_breast': 'left',
   'clock_position': 'n/a',
   'distance_from_nipple': 'n/a'\},
  'depth': 'n/a',
  'anatomical_region': 'n/a',
  'type': 'seroma',
  'shape': 'n/a',
  'orientation': 'n/a',
  'margin': 'n/a',
  'echogenicity': 'n/a',
  'calcifications': 'n/a',
  'vascularity': 'n/a',
  'posterior_features': 'n/a',
  'suspicion': 'n/a',
  'subtype': 'n/a',
  'next_step': 'n/a'\},
 \{'location': \{'side_of_breast': 'left',
   'clock_position': '3',
   'distance_from_nipple': '9'\},
  'depth': 'n/a',
  'anatomical_region': 'n/a',
  'type': 'cyst',
  'shape': 'n/a',
  'orientation': 'n/a',
  'margin': 'n/a',
  'echogenicity': 'n/a',
  'calcifications': 'n/a',
  'vascularity': 'n/a',
  'posterior_features': 'n/a',
  'suspicion': 'n/a',
  'subtype': 'n/a',
  'next_step': 'n/a'\}]
\end{MyVerbatim}
\noindent 
\noindent
\textbf{2. Lesion Attribute Confusion}

\noindent In the following example, the model incorrectly attributes "6 month follow up," which is associated to the lesion at 12 o'clock, to the lesion at 1 o'clock.
\begin{MyVerbatim}
Observation: RIGHT: There is a stable from [DATE] 1 cm benign mass right breast at 12 o'clock. Additionally, there is a stable from [DATE] 0.7 cm mass right breast at 1 o'clock. 

Impression: PROBABLY BENIGN - FOLLOW-UP RECOMMENDED A follow-up ultrasound in 6 months is recommended for probably benign mass at right 12:00 to document.  12 month follow-up is recommended for additional probably benign mass described above at right 1:00 to demonstrate stability.


\textbf{Prediction:}
[\{'location': \{'side_of_breast': 'right',
   'clock_position': '12',
   'distance_from_nipple': 'n/a'\},
  'depth': 'n/a',
  'anatomical_region': 'n/a',
  'type': 'mass',
  'shape': 'n/a',
  'orientation': 'n/a',
  'margin': 'n/a',
  'echogenicity': 'n/a',
  'calcifications': 'n/a',
  'vascularity': 'n/a',
  'posterior_features': 'n/a',
  'suspicion': 'n/a',
  'subtype': 'n/a',
  'next_step': '6 month follow-up'\},
 \{'location': \{'side_of_breast': 'right',
   'clock_position': '1',
   'distance_from_nipple': 'n/a'\},
  'depth': 'n/a',
  'anatomical_region': 'n/a',
  'type': 'mass',
  'shape': 'n/a',
  'orientation': 'n/a',
  'margin': 'n/a',
  'echogenicity': 'n/a',
  'calcifications': 'n/a',
  'vascularity': 'n/a',
  'posterior_features': 'n/a',
  'suspicion': 'n/a',
  'subtype': 'n/a',
  'next_step': '6 month follow-up'\}]

\textbf{Ground Truth:}
[\{'location': \{'side_of_breast': 'right',
   'clock_position': '12',
   'distance_from_nipple': 'n/a'\},
  'depth': 'n/a',
  'anatomical_region': 'n/a',
  'type': 'mass',
  'shape': 'n/a',
  'orientation': 'n/a',
  'margin': 'n/a',
  'echogenicity': 'n/a',
  'calcifications': 'n/a',
  'vascularity': 'n/a',
  'posterior_features': 'n/a',
  'suspicion': 'probably benign',
  'subtype': 'n/a',
  'next_step': '6 month follow-up'\},
 \{'location': \{'side_of_breast': 'right',
   'clock_position': '1',
   'distance_from_nipple': 'n/a'\},
  'depth': 'n/a',
  'anatomical_region': 'n/a',
  'type': 'mass',
  'shape': 'n/a',
  'orientation': 'n/a',
  'margin': 'n/a',
  'echogenicity': 'n/a',
  'calcifications': 'n/a',
  'vascularity': 'n/a',
  'posterior_features': 'n/a',
  'suspicion': 'probably benign',
  'subtype': 'n/a',
  'next_step': '12 month follow-up'\}
\end{MyVerbatim}

\noindent \textbf{3. Handling "N/A"}

\noindent Please refer to the example in \textbf{Lesion Attribute Confusion}. The model fails to predict the actual value for suspicion of malignancy, resulting in "N/A".

\section{Percentage of N/A Value in each key}
\label{appendix-g}
\begin{figure}[h]
    \centering
    \includegraphics[width=0.7\textwidth]{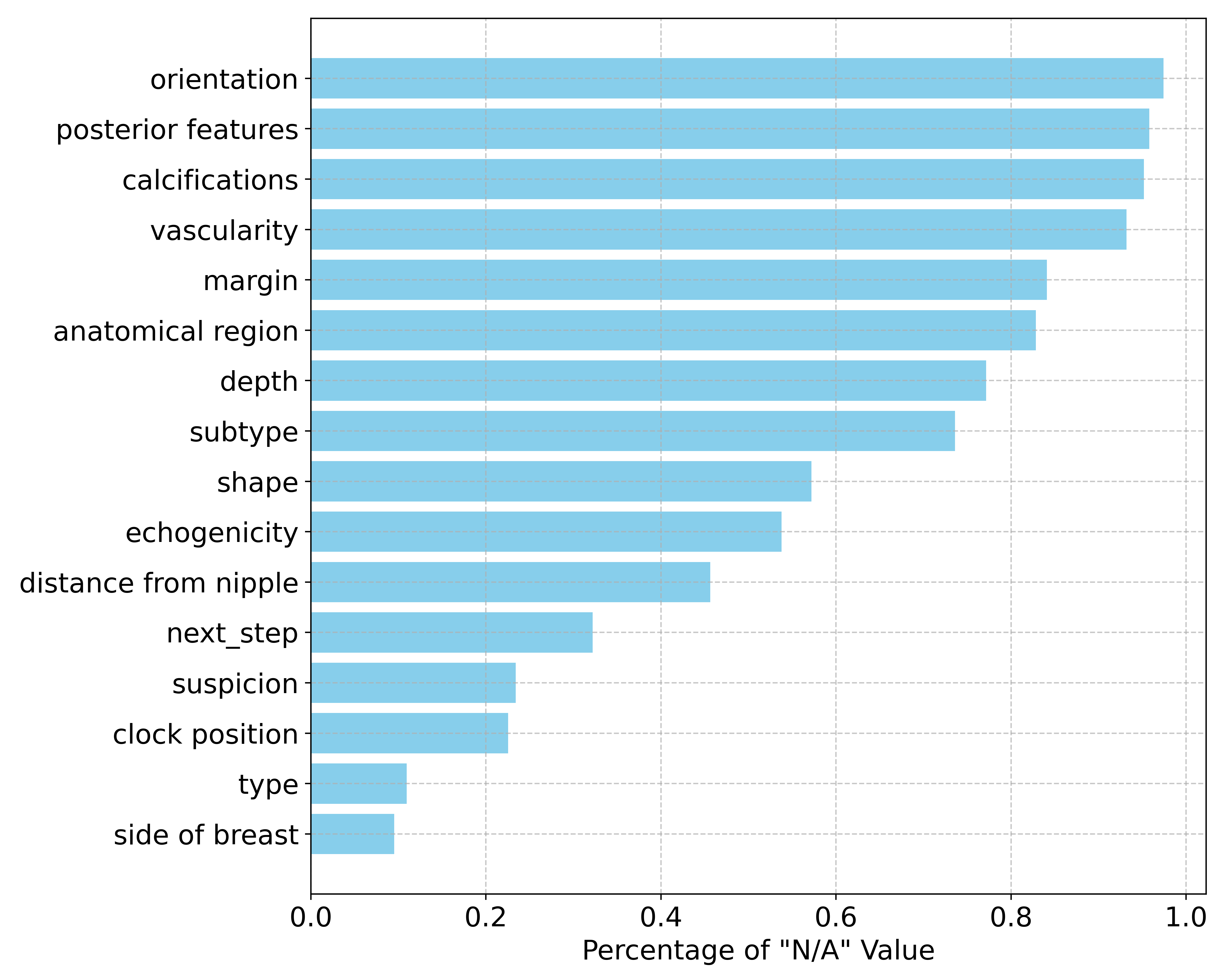}
    \label{fig:example}
\end{figure}

\end{document}